\documentclass[conference]{IEEEtran}

\newcommand{\RN}[1]{\uppercase\expandafter{\romannumeral#1}}

\usepackage[american]{babel}

\usepackage{pgfplots}
\usepackage{bm}
\usepackage{enumitem}

\usepackage{tikz,pgfplots,filecontents}
\usetikzlibrary{arrows,positioning,shapes,automata,fit,calc,decorations.pathreplacing,angles,quotes,backgrounds}
\usetikzlibrary{spy}

\usepackage[bookmarks=false]{hyperref}

\makeatletter
\@ifpackageloaded{babel}%
    {%
       \addto\extrasamerican{%
                }%
       \addto\extrasngerman{%
                }%
    }{\relax}
\makeatother

\ifCLASSOPTIONcompsoc
  \usepackage[nocompress]{cite}
\else
  \usepackage{cite}
\fi
\usepackage{pgfplots}
\usepackage{tikz}
\usetikzlibrary{arrows,positioning,shapes,automata,fit,calc,decorations.pathreplacing,angles,quotes,backgrounds}
\usetikzlibrary{spy}
\definecolor{tucol1}{rgb}{1.0 0.5 0.05}
\definecolor{tucol2}{rgb}{0.52,0.72,0.10}
\definecolor{tucol3}{rgb}{0.0 0.0 0.8}
\definecolor{tucol5}{rgb}{0.8 0.8 0}
\definecolor{tucol4}{rgb}{0.8 0 0.8}
\definecolor{tucol6}{rgb}{0 0.8 0.8}

\ifCLASSINFOpdf
\else
\fi
\usepackage{amsmath}
\usepackage{array}
\usepackage{fixltx2e}
\usepackage{dblfloatfix}

\usepackage{url}

\usepackage{csquotes}

\usepackage{booktabs}
\usepackage{tabularx}
\usepackage{multirow}
\newcolumntype{L}[1]{>{\hsize=#1\hsize\raggedright\arraybackslash}X}
\newcolumntype{R}[1]{>{\hsize=#1\hsize\raggedleft\arraybackslash}X}
\newcolumntype{C}[1]{>{\hsize=#1\hsize\centering\arraybackslash}X}

\usepackage{tcolorbox}

\usepackage[style=long,
            nopostdot,
            acronym,
            ]{glossaries}            
\glsdisablehyper
\makeglossaries

\newacronym{adam}{Adam}{Adaptive Moment Estimation}
\newacronym{bcel}{BCEL}{Binary Cross Entropy Loss}
\newacronym{cnn}{CNN}{Convolutional Neural Network}
\newacronym{gw}{GW}{George Washington Database}
\newacronym{iam}{IAM-DB}{IAM Handwriting Database}
\newacronym{mlp}{MLP}{Multilayer Perceptron}
\newacronym{phoc}{PHOC}{Pyramidal Histogram of Characters}
\newacronym{qbe}{QbE}{Query-by-Example}
\newacronym{qbs}{QbS}{Query-by-String}
\newacronym{relu}{ReLU}{Rectified Linear Units}
\newacronym{resnet}{ResNet}{Residual Network}
\newacronym{densenet}{DenseNet}{Densely Connected Convolutional Network}
\newacronym{tpp}{TPP}{Temporal Pyramid Pooling}

\newglossaryentry{tppnet}{name=TPP-PHOCNet, description={}}
\newglossaryentry{phoclenet}{name=PHOCLeNet, description={}}
\newglossaryentry{phocresnet}{name=PHOCResNet, description={}}
\newglossaryentry{phocdensenet}{name=PHOCDenseNet, description={}}

\begin{document}
\begin{NoHyper}
\title{Exploring Architectures\\for CNN-Based Word Spotting}

\author{
\IEEEauthorblockN{Eugen Rusakov, Sebastian Sudholt, Fabian Wolf, Gernot A. Fink}
\IEEEauthorblockA{	Department of Computer Science\\
					TU Dortmund University\\
					44221 Dortmund, Germany\\
					Email:\{eugen.rusakov, sebastian.sudholt, fabian2.wolf, gernot.fink\}@tu-dortmund.de}
}
\maketitle

\begin{abstract}
The goal in word spotting is to retrieve parts of document images which are relevant with respect to a certain user-defined query.
The recent past has seen attribute-based \acrlongpl{cnn} take over this field of research.
As is common for other fields of computer vision, the \acrshortpl{cnn} used for this task are already considerably deep.
The question that arises, however, is:
How complex does a \acrshort{cnn} have to be for word spotting?
Are increasingly deeper models giving increasingly better results or does performance behave asymptotically for these architectures?
On the other hand, can similar results be obtained with a much smaller \acrshort{cnn}?
The goal of this paper is to give an answer to these questions.
Therefore, the recently successful \gls{tppnet} will be compared to a \acrlong{resnet}, a \acrlong{densenet} and a LeNet architecture empirically.
As will be seen in the evaluation, a complex model can be beneficial for word spotting on harder tasks such as the IAM Offline Database but gives no advantage for \enquote{easier} benchmarks such as the George Washington Database.
\end{abstract}


%

\section{Introduction}\label{sec:introduction}
Word Spotting in handwritten documents refers to the image retrieval task with the goal 
to obtain a list of word images based on a relevance ranking for a given user's query. 
Since the introduction of word spotting in handwritten documents by Manmatha et. al. \cite{Manmatha96}, 
many promising approaches were proposed. Over the years the performance achieved by 
different approaches steadily increased mainly by introducing more powerful feature representations. 

In \cite{LeCun1998-GBL} a \emph{\gls{cnn}}, trained with the 
backpropagation algorithm, was successfully applied on a recognition task consisting of 
handwritten digits. Encouraged by the great success of \glspl{cnn} in recognition tasks 
\cite{Krizhevsky2012-ICW}  \cite{Simonyan2015-VDC}, Sudholt et. al. \cite{Sudholt17} 
proposed the \gls{tppnet}, a \gls{cnn} based on the attribute representation 
proposed by Almazan et. al \cite{Almazan14} namely the \emph{\gls{phoc}}. 
In 2015 He et. al. \cite{He16}  introduced the \gls{resnet} 
and achvied state-of-the-art results in image classification. 
Despite of an enormous depth the \glspl{resnet} can be efficiently trained by 
proposing residual modules containing skip-connections. Only one year later 
the \gls{densenet} \cite{Huang2017} was introduced with an enhanced approach concerning 
skip-connections. 
In contrast to \glspl{resnet}, which are based on the summation of layer output and skip-connection, the \gls{densenet} is built upon the respective concatenation.
This results in the connection of every layer to all subsequent layers and hence the name ''densely connected''.


In this work, we explore the \gls{tppnet}, \gls{resnet}, and \gls{densenet} on three standard 
word spotting benchmarks. All three architectures were adjusted and specifically designed 
for word spotting, by using the \gls{phoc} attribute representation. We evaluate the \glspl{cnn} 
for both the \gls{qbe} and \gls{qbs} scenario. This work shows that deeper networks as well as 
newer architectures do not necessarily perform better.

\autoref{sec:relatedwork} presents segmentation-based word spotting methods and gives a brief review of 
the explored CNN architectures.
The CNN architectures we used for this work are presented in \autoref{sec:method} followed by the evaluation in \autoref{sec:evaluation}. 
Finally a conclusion is drawn over the proposed exploration in \autoref{sec:conclusion}.


\section{Related Work}\label{sec:relatedwork}
\subsection{Word Spotting}\label{sec:rel_work_wordspotting}

The goal in word spotting is to retrieve word images from a given 
document image collection, rank these word images by a certain 
criteria, and sort them by their relevance with respect to 
a certain query. These queries can either be word images, 
defined by a user cropping a snipped from a document page, 
or defining a word string which needs to be retrieved.
Based on these two query definitions, word spotting distilled 
two scenarios namely \gls{qbe} and \gls{qbs}.
For \gls{qbe}, queries are given as word images and retrieval is 
based on a distance-based comparison. To perform such a 
comparison, a mapping needs to be find from a 
visual to a vector representation.
In the \gls{qbs} scenario, queries are given as word strings. 
In contrast to \gls{qbe}, the user is not required to search 
for a visual example of a word in a document collection.

Next to several approaches to address the problem of query representations 
based on visual similarity \cite{Rath07}, \cite{Rusinol15}. A 
very influential approach for learning a mapping from a visual to textual 
representation was presented in \cite{Almazan14}. 
The authors proposed a word string embedding in order to project word 
images into a common space based on a binary histogram of 
character occurrences namely \emph{Pyramidal Histogram of Characters \gls{phoc}}. A given word string is partitioned based on pyramidal 
levels, where each level splits the string with respect to the 
level number. For example, level one do not splits the string, 
whereas level two splits the string in two halves 
(ex. "home" into "ho" and "me"), level three splits the string in 
three parts, and so on.

Due to the great success of the \gls{phoc} word embedding, the focus 
concerning feature representation shifted towards trainable feature extractions. The \gls{phoc} embedding was originally proposed using 
low-level image descriptors like SIFT \cite{lowe04} in combination 
with Fisher-Vectors \cite{Perronnin10} and a linear SVM as classifier. 
Inspired by an outstanding performance of \glspl{cnn} in many 
computer vision classification tasks \cite{Krizhevsky2012-ICW},   \cite{Simonyan2015-VDC}, Sudholt et al proposed the \gls{tppnet} 
with the idea to replace the classification pipeline in \cite{Almazan14} 
with jointly trainable \glspl{cnn}.

\subsection{Deep Learning}

Convolutional Neural Networks (CNNs) have made their mark in 
various fields of computer vision and become the dominant pattern 
recognition approach for visual object recognition. Introduced 
over 20 years ago \cite{LeCun1989}, \glspl{cnn} were able 
to achieve comparable results but quickly comes to their limits. 
Based on the usage of a sigmoidal activation function, \glspl{cnn} 
were limited to have deep architectures exploiting more convolutional 
layers. The key contribution was made by introducing the 
\emph{Rectified Linear Unit (ReLU)} as activation function \cite{Glorot11}. 
Since that \glspl{cnn} were designed to have deeper architectures, starting 
with the AlexNet \cite{Krizhevsky2012-ICW} using 5 convolutional layers. 
In 2015 the VGGNet \cite{Simonyan2015-VDC} was proposed using 19 
convolutional layers. One year later the first two networks were 
capable of surpassing the 100-layer barrier, namely the Highway Networks \cite{Srivastava15} 
and the Residual Networks (ResNets) \cite{He16}. 
A \emph{degradation} problem has been exposed when deeper networks are able 
to start converging. The accuracy gets saturated with the increasing network depth
and than rapidly degrades. Adding more layers to the network leads to higher 
training error. Against the first intuition this degradation is not caused 
by overfitting as reported in \cite{He14} and \cite{Srivastava15}.

\subsubsection{ResNet}\label{sec:rel_work_resnet}
The \emph{Residual Networks (ResNets)} address the problem of degradation, using 
so-called \emph{Residual Blocks} in the network architecture. The authors let 
the layers explicitly fit a residual mapping \cite{He16}, formally denoting 
the mapping as $\mathcal{H}(x) = \mathcal{F}(x) + x$. Here $x$ is the input 
of a residual building block and $\mathcal{F}(x)$ the output of the layers within 
a building block. $\mathcal{H}(x)$ is than the summation of both in order to 
obtain the mapping, with the identity $x$ as a "shortcut connection". The 
ResNet comes with two versions of residual building blocks. For "shallower" 
residual networks ($18$- or $34$-layers) the building blocks contain two $3 \times 3$ convolution 
layers. Whereas the "deeper" networks ($50$-, $101$-, and $151$-layers) are built on 
so-called residual bottleneck blocks, containing $1 \times 1$, $3 \times 3$, and $1 \times 1$ 
convolution layers consecutive. As the first $1 \times 1$ layer reduces the dimensionality, 
the second $1 \times 1$ convolution layer is restoring the dimensionality. 
Each residual bottleneck block has an ordinal $n$ which determines the number of filter kernels used in a specific block as follows:
The first and second layer in the block have $n/4$ filters while the third layer makes use of $n$ filters. 
For example if $n=256$ than layer one and two have $64$ filter kernels and layer three has $256$ filter kernels.

\subsubsection{DenseNets}\label{sec:rel_work_densenet}

Similar to the \gls{resnet} architecture,  \glspl{densenet} \cite{Huang2017} also utilize identity connections.
A skip-connection bypasses each layer.
This identity path is then concatenated with the feature-maps generated by the layer.
Following this connectivity scheme, each layer receives all preceding feature-maps as an input.
The number of feature-maps added by each layer is referred to as the network's growth rate.
Already small growth rates are sufficient to achieve state-of-the-art results, resulting in very narrow layers.
To avoid the concatenation of differently sized feature-maps, \glspl{densenet} are organized into densely connected blocks.
Consequently, pooling layers are only used outside the dense blocks.
The model compactness of \glspl{densenet} is further improved by the introduction of compression layers.
A compression layer corresponds to a convolutional layer with kernel size one.
\glspl{densenet} tend to make stronger use of high-level features learned at the end of a dense block.
Therefore, the convolutional layer reduces the number of features maps by a compression factor.
The \gls{densenet} architecture outperforms other networks such as \glspl{resnet} in various state-of-the-art benchmark experiments in the field of image classification.
It has been shown, that the dense architecture is especially parameter efficient and achieves competitive results, with a significantly less numbers of parameters.
\section{Architectures for Word Spotting}\label{sec:method}
In this section, we describe the different architectures we evaluated for the word spotting task.
The order of the different architectures is chosen such that the networks get increasingly deeper and more complex from an architectural point of view.
All presented architectures are trained to predict the \gls{phoc} representation~\cite{Almazan14} using the \gls{bcel}.
While there exist other word string embedding types for word spotting, this combination achieves consistently good results for a number of different word spotting benchmarks and has the fastest training times~\cite{Sudholt2018-ACF}.

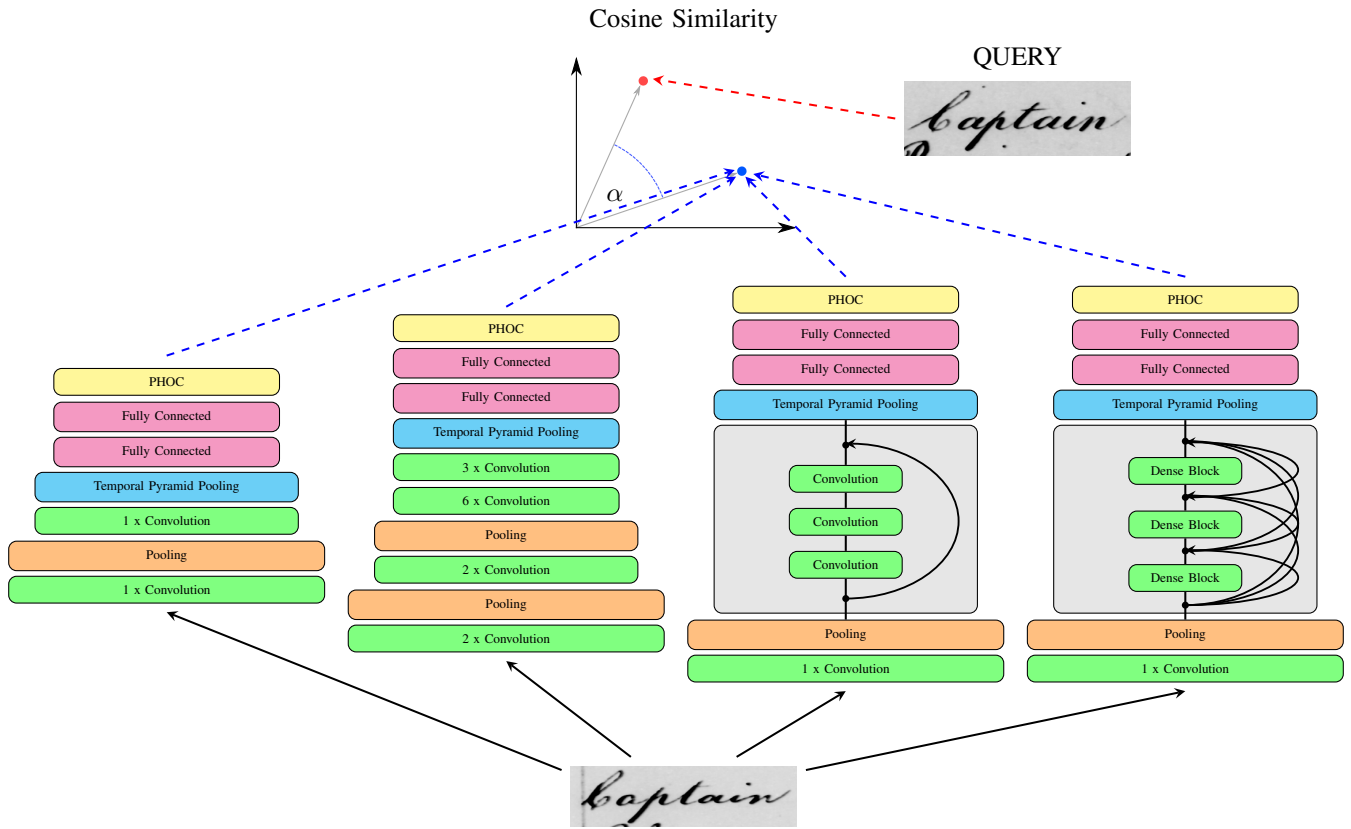
\begin{figure*}
	\centering
	\begin{tikzpicture}[
	node distance = 1cm,
	auto,
	scale=1.0,
	transform shape,
	triangle/.style = {regular polygon, regular polygon sides=3 },
	node rotated/.style = {rotate=180},
	border rotated/.style = {shape border rotate=180}
	]
	
	\node[ 
	line width=0.2mm,
	rounded corners=10pt, 
	minimum height=9cm, 
	minimum width=0.95\textwidth]
	at (0,0) 
	(word_hypotheses)
	{};
	
	\node[
	line width=0.2mm,
	rounded corners=2pt, 
	minimum height=0cm, 
	minimum width=0cm]
	at ([xshift=0em] word_hypotheses.north)
	(cosine)
	{
		\includegraphics[width=3cm]{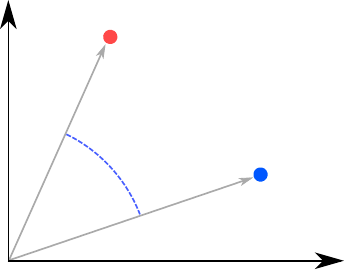}
	};
	\node at ([xshift=-2.6em, yshift=-1.9em] word_hypotheses.north) {$\alpha$};
	
	\node[
	line width=0.2mm,
	rounded corners=2pt, 
	minimum height=3cm, 
	minimum width=3cm]
	at ([yshift=1em, xshift=0em] cosine.north)
	(query_label)
	{
		\normalsize Cosine Similarity
	};

	\node[
	line width=0.2mm,
	rounded corners=2pt, 
	minimum height=3cm, 
	minimum width=3cm]
	at ([yshift=1em, xshift=8em] cosine.east)
	(query_captain)
	{
		\includegraphics[width=3cm]{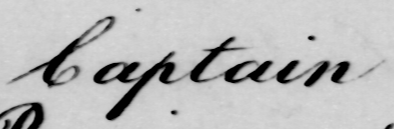}
	};

	\node[
	line width=0.2mm,
	rounded corners=2pt, 
	minimum height=3cm, 
	minimum width=3cm]
	at ([yshift=-2em, xshift=0em] query_captain.north)
	(query_label)
	{
		\normalsize QUERY
	};
	
	\node[ 
	rounded corners=3pt,
	minimum height=3.4cm, 
	minimum width=4.5cm]
	at ([yshift=0em, xshift=1.75cm] word_hypotheses.west)
	(lenet)
	{
	};
	\node[
	rounded corners=3pt,
	rectangle, draw,
	minimum height=0.0cm, 
	minimum width=4.2cm,
	fill=green!50]
	(conv1)
	at ([yshift=0.3cm, xshift=0.0cm] lenet.south)
	{
		\tiny{1 x Convolution}	
	};
	\node[
	rounded corners=3pt,
	rectangle, draw,
	minimum height=0.0cm, 
	minimum width=4.2cm,
	fill=orange!50]
	(pool1)
	at ([yshift=0.75em, xshift=0cm] conv1.north)
	{
		\tiny{Pooling}	
	};
	\node[
	rounded corners=3pt,
	rectangle, draw,
	minimum height=0.0cm, 
	minimum width=3.5cm,
	fill=green!50]
	(conv2)
	at ([yshift=0.75em, xshift=0cm] pool1.north)
	{
		\tiny{1 x Convolution}	
	};
	\node[
	rounded corners=3pt,
	rectangle, draw,
	minimum height=0.0cm, 
	minimum width=3.5cm,
	fill=cyan!50]
	(tpp)
	at ([yshift=0.75em, xshift=0cm] conv2.north)
	{
		\tiny{Temporal Pyramid Pooling}	
	};
	\node[
	rounded corners=3pt,
	rectangle, draw,
	minimum height=0.0cm, 
	minimum width=3.0cm,
	fill=magenta!50]
	(fc1)
	at ([yshift=0.75em, xshift=0cm] tpp.north)
	{
		\tiny{Fully Connected}	
	};
	\node[
	rounded corners=3pt,
	rectangle, draw,
	minimum height=0.0cm, 
	minimum width=3.0cm,
	fill=magenta!50]
	(fc2)
	at ([yshift=0.75em, xshift=0cm] fc1.north)
	{
		\tiny{Fully Connected}	
	};
	\node[
	rounded corners=3pt,
	rectangle, draw,
	minimum height=0.0cm, 
	minimum width=3.0cm,
	fill=yellow!50]
	(phoc_tpp)
	at ([yshift=0.75em, xshift=0cm] fc2.north)
	{
		\tiny{PHOC}	
	};
	
	\node[ 
	right = 0.0cm of lenet,
	rounded corners=3pt,
	minimum height=4.7cm, 
	minimum width=4.5cm]
	(tppnet)
	{
	};
	\node[
	rounded corners=3pt,
	rectangle, draw,
	minimum height=0.0cm, 
	minimum width=4.2cm,
	fill=green!50]
	(conv1)
	at ([yshift=0.3cm, xshift=0cm] tppnet.south)
	{
		\tiny{2 x Convolution}	
	};
	\node[
	rounded corners=3pt,
	rectangle, draw,
	minimum height=0.0cm, 
	minimum width=4.2cm,
	fill=orange!50]
	(pool1)
	at ([yshift=0.75em, xshift=0cm] conv1.north)
	{
		\tiny{Pooling}	
	};
	\node[
	rounded corners=3pt,
	rectangle, draw,
	minimum height=0.0cm, 
	minimum width=3.5cm,
	fill=green!50]
	(conv2)
	at ([yshift=0.75em, xshift=0cm] pool1.north)
	{
		\tiny{2 x Convolution}	
	};
	\node[
	rounded corners=3pt,
	rectangle, draw,
	minimum height=0.0cm, 
	minimum width=3.5cm,
	fill=orange!50]
	(pool2)
	at ([yshift=0.75em, xshift=0cm] conv2.north)
	{
		\tiny{Pooling}	
	};
	\node[
	rounded corners=3pt,
	rectangle, draw,
	minimum height=0.0cm, 
	minimum width=3.0cm,
	fill=green!50]
	(conv3)
	at ([yshift=0.75em, xshift=0cm] pool2.north)
	{
		\tiny{6 x Convolution}	
	};
	\node[
	rounded corners=3pt,
	rectangle, draw,
	minimum height=0.0cm, 
	minimum width=3.0cm,
	fill=green!50]
	(conv4)
	at ([yshift=0.75em, xshift=0cm] conv3.north)
	{
		\tiny{3 x Convolution}	
	};
	\node[
	rounded corners=3pt,
	rectangle, draw,
	minimum height=0.0cm, 
	minimum width=3.0cm,
	fill=cyan!50]
	(tpp)
	at ([yshift=0.75em, xshift=0cm] conv4.north)
	{
		\tiny{Temporal Pyramid Pooling}	
	};
	\node[
	rounded corners=3pt,
	rectangle, draw,
	minimum height=0.0cm, 
	minimum width=3.0cm,
	fill=magenta!50]
	(fc1)
	at ([yshift=0.75em, xshift=0cm] tpp.north)
	{
		\tiny{Fully Connected}	
	};
	\node[
	rounded corners=3pt,
	rectangle, draw,
	minimum height=0.0cm, 
	minimum width=3.0cm,
	fill=magenta!50]
	(fc2)
	at ([yshift=0.75em, xshift=0cm] fc1.north)
	{
		\tiny{Fully Connected}	
	};
	\node[
	rounded corners=3pt,
	rectangle, draw,
	minimum height=0.0cm, 
	minimum width=3.0cm,
	fill=yellow!50]
	(phoc_tpp)
	at ([yshift=0.75em, xshift=0cm] fc2.north)
	{
		\tiny{PHOC}	
	};
	
	
	\node[ 
	right = 0.0cm of tppnet,
	rounded corners=3pt,
	minimum height=5.5cm, 
	minimum width=4.5cm]
	(resnet)
	{
		\Large{PHOCResNet}
	};
	\node[
	rounded corners=3pt,
	rectangle, draw,
	minimum height=0.0cm, 
	minimum width=4.2cm,
	fill=green!50]
	(conv1)
	at ([yshift=0.3cm, xshift=0.0cm] resnet.south)
	{
		\tiny{1 x Convolution}	
	};
	\node[
	rounded corners=3pt,
	rectangle, draw,
	minimum height=0.0cm, 
	minimum width=4.2cm,
	fill=orange!50]
	(pool1)
	at ([yshift=0.75em, xshift=0cm] conv1.north)
	{
		\tiny{Pooling}	
	};
	\node[
	rounded corners=3pt,
	rectangle, draw,
	minimum height=2.5cm, 
	minimum width=3.5cm,
	fill=black!10]
	(resnetblock)
	at ([yshift=3.8em, xshift=0cm] pool1.north)
	{
	};
	\node[
	rounded corners=3pt,
	rectangle, draw,
	minimum height=0.0cm, 
	minimum width=1.5cm,
	fill=green!50]
	(block_conv1)
	at ([yshift=1.85em, xshift=0cm] resnetblock.south)
	{
		\tiny{Convolution}	
	};
	\node[
	rounded corners=3pt,
	rectangle, draw,
	minimum height=0.0cm, 
	minimum width=1.5cm,
	fill=green!50]
	(block_conv2)
	at ([yshift=1.1em, xshift=0cm] block_conv1.north)
	{
		\tiny{Convolution}	
	};
	\node[
	rounded corners=3pt,
	rectangle, draw,
	minimum height=0.0cm, 
	minimum width=1.5cm,
	fill=green!50]
	(block_conv3)
	at ([yshift=1.1em, xshift=0cm] block_conv2.north)
	{
		\tiny{Convolution}	
	};
	
	\node[
	minimum height=-1.0cm, 
	minimum width=-1.0cm]
	(start)
	at ([yshift=-0.75em, xshift=0cm] block_conv1.south)
	{
			\Huge{.}
	};
	
	\node[
	minimum height=-1.0cm, 
	minimum width=-1.0cm]
	(end)
	at ([yshift=0.75em, xshift=0cm] block_conv3.north)
	{
			\Huge{.}
	};
	
	\node[
	rounded corners=3pt,
	rectangle, draw,
	minimum height=0.0cm, 
	minimum width=3.5cm,
	fill=cyan!50]
	(tpp_resnet)
	at ([yshift=0.75em, xshift=0cm] resnetblock.north)
	{
		\tiny{Temporal Pyramid Pooling}	
	};
	\node[
	rounded corners=3pt,
	rectangle, draw,
	minimum height=0.0cm, 
	minimum width=3.0cm,
	fill=magenta!50]
	(fc1)
	at ([yshift=0.75em, xshift=0cm] tpp_resnet.north)
	{
		\tiny{Fully Connected}	
	};
	\node[
	rounded corners=3pt,
	rectangle, draw,
	minimum height=0.0cm, 
	minimum width=3.0cm,
	fill=magenta!50]
	(fc2)
	at ([yshift=0.75em, xshift=0cm] fc1.north)
	{
		\tiny{Fully Connected}	
	};
	\node[
	rounded corners=3pt,
	rectangle, draw,
	minimum height=0.0cm, 
	minimum width=3.0cm,
	fill=yellow!50]
	(phoc_tpp)
	at ([yshift=0.75em, xshift=0cm] fc2.north)
	{
		\tiny{PHOC}	
	};

	\draw[line width=0.25mm, -stealth] (start) arc [x radius = 1.5cm, y radius = 1.03cm, start angle = -90,end angle = +90] (end);
	
	\draw[line width=0.27mm] (pool1) -- (block_conv1);
	
	\draw[line width=0.27mm] (block_conv1) -- (block_conv2);
	
	\draw[line width=0.27mm] (block_conv2) -- (block_conv3);
	
	\draw[line width=0.27mm] (block_conv3) -- (tpp_resnet);
	

	\node[ 
	right = 0.0cm of resnet,
	rounded corners=3pt,
	minimum height=5.5cm, 
	minimum width=4.5cm]
	(densenet)
	{
	};
	\node[
	rounded corners=3pt,
	rectangle, draw,
	minimum height=0.0cm, 
	minimum width=4.2cm,
	fill=green!50]
	(dense_conv1)
	at ([yshift=0.3cm, xshift=0.0cm] densenet.south)
	{
		\tiny{1 x Convolution}	
	};
	\node[
	rounded corners=3pt,
	rectangle, draw,
	minimum height=0.0cm, 
	minimum width=4.2cm,
	fill=orange!50]
	(dense_pool1)
	at ([yshift=0.75em, xshift=0cm] dense_conv1.north)
	{
		\tiny{Pooling}	
	};
	\node[
	rounded corners=3pt,
	rectangle, draw,
	minimum height=2.5cm, 
	minimum width=3.5cm,
	fill=black!10]
	(denseblock)
	at ([yshift=3.8em, xshift=0cm] dense_pool1.north)
	{
	};
	\node[
	rounded corners=3pt,
	rectangle, draw,
	minimum height=0.0cm, 
	minimum width=1.5cm,
	fill=green!50]
	(denseblock_conv1)
	at ([yshift=1.35em, xshift=0cm] denseblock.south)
	{
		\tiny{Dense Block}	
	};
	\node[
	rounded corners=3pt,
	rectangle, draw,
	minimum height=0.0cm, 
	minimum width=1.5cm,
	fill=green!50]
	(denseblock_conv2)
	at ([yshift=1.5em, xshift=0cm] denseblock_conv1.north)
	{
		\tiny{Dense Block}	
	};
	\node[
	rounded corners=3pt,
	rectangle, draw,
	minimum height=0.0cm, 
	minimum width=1.5cm,
	fill=green!50]
	(denseblock_conv3)
	at ([yshift=1.5em, xshift=0cm] denseblock_conv2.north)
	{
		\tiny{Dense Block}	
	};
	
	\node[
	minimum height=-1.0cm, 
	minimum width=-1.0cm]
	(dense_start)
	at ([yshift=-0.5em, xshift=0cm] denseblock_conv1.south)
	{
		\Huge{.}
	};
	
	\node[
	minimum height=-1.0cm, 
	minimum width=-1.0cm]
	(dense_node_1)
	at ([yshift=0.5em, xshift=0cm] denseblock_conv1.north)
	{
		\Huge{.}
	};
	
	\node[
	minimum height=-1.0cm, 
	minimum width=-1.0cm]
	(dense_node_2)
	at ([yshift=0.5em, xshift=0cm] denseblock_conv2.north)
	{
		\Huge{.}
	};
	
	\node[
	minimum height=-1.0cm, 
	minimum width=-1.0cm]
	(dense_end)
	at ([yshift=0.6em, xshift=0cm] denseblock_conv3.north)
	{
		\Huge{.}
	};
	
	\node[
	rounded corners=3pt,
	rectangle, draw,
	minimum height=0.0cm, 
	minimum width=3.5cm,
	fill=cyan!50]
	(tpp_densenet)
	at ([yshift=0.75em, xshift=0cm] denseblock.north)
	{
		\tiny{Temporal Pyramid Pooling}	
	};
	\node[
	rounded corners=3pt,
	rectangle, draw,
	minimum height=0.0cm, 
	minimum width=3.0cm,
	fill=magenta!50]
	(dense_fc1)
	at ([yshift=0.75em, xshift=0cm] tpp_densenet.north)
	{
		\tiny{Fully Connected}	
	};
	\node[
	rounded corners=3pt,
	rectangle, draw,
	minimum height=0.0cm, 
	minimum width=3.0cm,
	fill=magenta!50]
	(dense_fc2)
	at ([yshift=0.75em, xshift=0cm] dense_fc1.north)
	{
		\tiny{Fully Connected}	
	};
	\node[
	rounded corners=3pt,
	rectangle, draw,
	minimum height=0.0cm, 
	minimum width=3.0cm,
	fill=yellow!50]
	(dense_phoc_tpp)
	at ([yshift=0.75em, xshift=0cm] dense_fc2.north)
	{
		\tiny{PHOC}	
	};

	\draw[line width=0.25mm, -stealth] (dense_start) arc [x radius = 1.5cm, y radius = 0.366cm, start angle = -90,end angle = +90] (dense_node_1);
	\draw[line width=0.25mm, -stealth] (dense_node_1) arc [x radius = 1.5cm, y radius = 0.366cm, start angle = -90,end angle = +90] (dense_node_2);
	\draw[line width=0.25mm, -stealth] (dense_node_2) arc [x radius = 1.5cm, y radius = 0.366cm, start angle = -90,end angle = +90] (dense_end);
	
	\draw[line width=0.25mm] (dense_start) arc [x radius = 1.5cm, y radius = 0.73cm, start angle = -90,end angle = +90] (dense_node_2);
	\draw[line width=0.25mm] (dense_node_1) arc [x radius = 1.5cm, y radius = 0.73cm, start angle = -90,end angle = +90] (dense_node_2);

	\draw[line width=0.25mm] (dense_start) arc [x radius = 1.5cm, y radius = 1.09cm, start angle = -90,end angle = +90] (dense_node_2);
		
	\draw[line width=0.27mm] (dense_pool1) -- (denseblock_conv1);
	
	\draw[line width=0.27mm] (denseblock_conv1) -- (denseblock_conv2);
	
	\draw[line width=0.27mm] (denseblock_conv2) -- (denseblock_conv3);
	
	\draw[line width=0.27mm] (denseblock_conv3) -- (tpp_densenet);
	
	
	\node[ 
	rounded corners=3pt, 
	minimum height=1cm, 
	minimum width=1cm]
	at ([xshift=0em, yshift=1em] word_hypotheses.south)
	(captain)
	{
		\includegraphics[width=3cm]{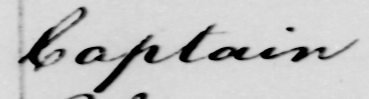}
	};

	\draw[line width=0.27mm, red, dashed, -stealth] (query_captain.west) -- ([yshift=2.5em, xshift=-5.8em] cosine.east);

	\draw[line width=0.27mm, -stealth] ([yshift=1em] captain.west) -- (lenet.south);
	
	\draw[line width=0.27mm, -stealth] ([xshift=-2em] captain.north) -- (tppnet.south);
	
	\draw[line width=0.27mm, -stealth] ([xshift=2em] captain.north) -- (resnet.south);
	
	\draw[line width=0.27mm, -stealth] ([yshift=1em] captain.east) -- (densenet.south);

	
	\draw[line width=0.27mm, blue!100, dashed, -stealth] (lenet.north) -- ([yshift=-0.95em, xshift=-2.7em] cosine.east);
	
	\draw[line width=0.27mm, blue!100, dashed, -stealth] (tppnet.north) -- ([yshift=-1.2em, xshift=-2.65em] cosine.east);
	
	\draw[line width=0.27mm, blue!100, dashed, -stealth] (resnet.north) -- ([yshift=-1.2em, xshift=-2.3em] cosine.east);
	
	\draw[line width=0.27mm, blue!100, dashed, -stealth] (densenet.north) -- ([yshift=-1.1em, xshift=-2.0em] cosine.east);
		
	
	\end{tikzpicture}
	\caption{Overview of the explored CNN architectures. The networks are from left to right, PHOCLeNet, TPP-PHOCNet, PHOCResNet, and PHOCDenseNet. The CNNs are trained to predict a desired word string embedding generated from the annotation. Word spotting can then be performed in the embedding space through a simple nearest neighbor search using the cosine similarity as distance metric. Please note, that due to space issues this overview is not an exactly architecture description especially for the residual bottleneck and dense blocks. 
	The exactly description is given in Sec. \ref{sec:method}.}
	\label{fig:architectures_overview}
\end{figure*}


\subsection{PHOCLeNet}
Although the title of first \gls{cnn} goes to the Neocognitron~\cite{Fukushima1980-NAS}, the \emph{LeNet} was the first \gls{cnn} to be trained in a supervised fashion with the backpropagation algorithm~\cite{LeCun1998-GBL}.
It consists of seven layers:
The first four layers build the convolutional part of the \gls{cnn} and are made up of two consecutive combinations of convolutional layer and max pooling layers.
While the first of these two layers uses filter $28 \times 28$ filter kernels in order to produce six feature maps, the second convolutional layer uses $14 \times 14$ kernels and produces $16$ feature maps.
After the convolutional part, the LeNet employs a \gls{mlp} with two hidden layers.
The number of the neurons in these two layers are $120$ and $84$.
The size of the output layer depends on the number of classes to be predicted from the \gls{cnn}.
All hidden layers of the LeNet which carry weights make use of a hyperbolic tangent as activation function.

For the purpose of predicting attributes with the LeNet, we make three subtle changes to the original architecture:
First, the hyperbolic tangent activation functions are replaced with \glspl{relu}.
The reason behind this is the vanishing gradient problem~(cf. e.g. \cite{Nielsen2015-NNA}):
Even though the LeNet architecture is shallow compared to current deep learning architectures, it still employs four layers which use the hyperbolic tangent as activation function.
Replacing them with \glspl{relu} eliminates the vanishing gradient problem as well as speeds up the training process.
The second change is replacing the last pooling layer with a \gls{tpp} layer~\cite{Sudholt17}.
This allows the \gls{cnn} to process images of different sizes.
The last change made to the architecture are the size of the convolutional and fully connected layers.
Here, we use the same changes as are done for the LeNet architecture in the Caffe deep learning library~\cite{Jia2014-CCA} and increase the trainable parameters of the \gls{cnn}:
The number of filters in the first convolutional layer is set to $20$ while for the second convolutional layer $50$ filters are used.
The hidden fully connected layers are enlarged to $500$ neurons.
While the network remains small from a parameter point of view compared to current architectures, it gains a considerable amount of power compared to the original version.

\subsection{TPP-PHOCNet}\label{sec:tppnet}
The \gls{tppnet}~\cite{Sudholt17} is an enhanced version of the PHOCNet~\cite{Sudholt16}.
Its architecture is inspired by the successful VGG16 \gls{cnn}~\cite{Simonyan2015-VDC}.
The convolutional part consists of $13$ convolutional layers with two max pooling layers inserted after the second and fourth convolutional layer.
As is done in the original VGG16, all layers make use of $3 \times 3$ filter kernels.
Compared to the LeNet and also the successful AlexNet~\cite{Krizhevsky2012-ICW}, this choice of filter sizes serves as regularization as it substantially decreases the amount of parameters in the convolutional layers.
While the original VGG16 also employed max pooling layers after the seventh and tenth convolutional layers, these max pooling layers are eliminated in the \gls{tppnet}.
The reason for this is that the word images used as input for the \gls{cnn} are usually much smaller than the natural images the VGG16 was designed for.
Removing them from the architecture allows for a smaller increase in receptive field for the later layers of the convolutional part.
After the convolutional layer, the \gls{tppnet} makes use of a \gls{tpp} layer in order to obtain a fixed image representation for images of varying sizes.
The \gls{mlp} part of the architecture consists of two hidden layers with $4096$ neurons each.
This is in accordance with the AlexNet and VGG16 \glspl{cnn}.
As these two layers carry the most trainable parameters and are prone to overfitting, a dropout of $50\%$ is applied to both hidden layers.
In dropout, the output of a given neuron is randomly set to $0$ during training.
The probability for a neuron to be dropped out is a meta-parameter which has to be defined by the user.
Using a dropout of $50\%$ is a popular practice when using hidden layers of these sizes in current \glspl{cnn} (cf. e.g. \cite{Krizhevsky2012-ICW,Simonyan2015-VDC,Zeiler2014-VAU,Sermanet2014-OIR}).

All convolutional and fully connected layers in the \gls{tppnet} use \glspl{relu} as activation function.

\subsection{PHOCResNet}
The PHOCResNet is a novel architecture we propose for word spotting.
It is inspired by the successful ResNet50 architecture~\cite{He16}.
This \gls{cnn} uses a $7\times7$ convolutional and a $3\times3$ max pooling layer in the beginning of the network which are then followed by $16$ residual bottleneck blocks.
The total number of convolutional layers in the \gls{cnn} is $49$.
The ordinal for the blocks are as follows:
Block 1 to 3 have ordinal $256$, blocks 4 to 7 have $512$, blocks 8 to 13 have $1024$ and the final three blocks have ordinal $2048$.
As the VGG16, the ResNet50 was designed for natural images as input.
In order to adapt it to attribute-based word spotting, we make four changes to the \gls{cnn}'s architecture:
The original ResNet50 architecture uses strided convolutions for down sampling after the blocks 3, 7 and 13 in addition to the down sampling obtained from the pooling layer.
As is done for the \gls{tppnet}, we reduce the number of down sampling operations due to the word images being comparably small in size.
Thus, we only use the first max pooling layer in addition to strided convolutions after the third residual block.
Second, we replace the average pooling used in the original ResNet50 with a \gls{tpp} layer in order to account for images of varying size.
The third change is made with respect to the fully connected part of the ResNet50.
Pre-experimental evaluations showed that learning a \gls{phoc} representation with the ResNet50 is quite difficult as it employs only a single fully connected layer in the end of the architecture.
We obtained substantially better results when replacing this single layer with an \gls{mlp} as was used in the \gls{tppnet}.
It seems, that the \gls{mlp} is more able to learn the \gls{phoc} representation of correlated binary values than  a single fully connected layer, i.e., a perceptron.
Thus, the same \gls{mlp} is used for the PHOCResNet as is done for the PHOCNet and \gls{tppnet} (cf. \autoref{sec:tppnet}).
The last change made in the PHOCResNet architecture with respect to the ResNet50 is that we delete all batch normalization layers.
The reason for this is of a technical nature:
We compute the gradients for mini-batches of varying sizes by obtaining the gradients for each image separately and then averaging them.
This means, that our effective batch size is $1$ and the weights are updated after performing backprop $b$ times where $b$ is the batch size.
The batch normalization layers would thus only be presented with mini-batches of size $1$ which would then be used for determining the mean and variance.
In addition, the feature maps obtained for the different mini batches have different sizes which would skew the computation of a global mean and variance value.
Hence, we elect to not use batch normalization in our network.

\subsection{PHOCDenseNet}
The PHOCDenseNet architecture is inspired by the the \gls{densenet} structure proposed in \cite{Huang2017}.
The network employs two densely connected blocks and a growth rate of $12$.
Pre-experimental evaluations showed, that different growth rates perform comparable well, while small growth rates has shown to be more parameter efficient.
Before entering the first block, a convolutional layer with $32$ output channels and a $2\times2$ average pooling layer are employed.
Following the dense connectivity pattern, the first block consists of $30$ convolutional layers with kernel sizes $3\times3$.
For the second block $60$ densely connected convolutional layers are used.
The transition layer between both blocks employ a convolutional layer with kernel size $1\times1$ and a $2\times2$ average pooling layer.
The convolutional layer compresses the number of feature maps by a factor of $0.5$
Analogue to the \textit{TPP-PHOCNet} architecture, the PHOCDenseNet makes use of a 5-level TPP layer in combination with a Multilayer Perceptron.

\section{Experiments}\label{sec:evaluation}
\begin{table*}[t]
	\centering
	\caption{Comparison of the different results obtained for the different experiments [\%].}
	\label{tab:results}
	\begin{tabularx}{0.8\textwidth}{L{2}C{0.8}C{0.8}C{0.8}C{0.8}C{0.8}C{0.8}C{0.8}C{0.8}C{0.8}C{0.8}}
		\toprule
		\multirow{2}{*}{Architecture/Method}	& \multicolumn{2}{c}{George Washington} && \multicolumn{2}{c}{IAM} && \multicolumn{2}{c}{Botany \textit{Train \RN{3}}} \\ 
		& \acrshort{qbe} & \acrshort{qbs} && \acrshort{qbe} & \acrshort{qbs} && \acrshort{qbe} & \acrshort{qbs}\\
		\midrule
		LeNet & $69.78$ & $80.68$ && $13.55$ & $27.86$ && $32.27$ & $21.82$ \\
		\gls{tppnet} & $97.29$ & $\textbf{98.47}$ && $85.03$ & $93.22$ && $92.89$ & $96.61$ \\
		PHOCResNet & $95.90$ & $97.91$ && $\textbf{88.35}$ & $\textbf{94.69}$ && $\textbf{95.92}$ & $\textbf{98.53}$ \\
		PHOCDenseNet & $95.20$ & $94.71$ && $72.06$ & $84.11$ && $70.26$ & $82.95$ \\[1ex]
		Deep Features~\cite{Krishnan16} & $94.41$ & $92.84$ && $84.24$ & $91.58$ && $-$ & $-$ \\
		Triplet-CNN~\cite{Wilkinson16}\footnotemark[1] & $\textbf{98.00}$ & $93.69$ && $81.58$ & $89.49$ && $-$ & $-$ \\
		AttributeSVM~\cite{Almazan14} & $93.04$ & $91.29$ && $55.73$ & $73.72$ && $-$ & $-$ \\
		\bottomrule
	\end{tabularx}
\end{table*}
For the experiments with the four architectures we used three benchmark datasets described in \autoref{sec:datasets} 
and a evaluation protocol (\autoref{sec:protocol}) for segmentation-based wordspotting commonly used in the literature. 
In \autoref{sec:training} we describe the training setup with all hyper-parameter used for training.
Afterwards we discuss the retrieval results achieved by our methods and compare the architectures in \autoref{sec:results}.
\subsection{Datasets}\label{sec:datasets}
We evaluate our method on three publicly available data sets. 
The first is the \textbf{George Washington (GW) data set}. 
It consists of $20$ pages that are containing $4\,860$ annotated words.
The pages originate from a letterbook and are quite homogeneous in their visual appearance.
However, particularly for smaller words the annotation is very sloppy. As the GW data set does
not have an official partitioning into training and test pages, we follow the common approach
and perform a four-fold cross validation. Thus, the data set is split into batches of five
consecutive documents each.

The second benchmark is the large \textbf{IAM off-line dataset} comprising $1\,539$ pages of 
modern handwritten English text containing $115\,320$ word images, written by $657$ different writers. 
We used the official partition available for writer independent text line recognition. We combined the training and 
validation set to $60\,453$ word images for training, and used the $13\,752$ word images in the test set for evaluation. 
We exclude the stop words as queries, as this is common in this benchmark.

\textbf{Botany in British India (Botany)} is the third benchmark introduced in the Handwritting 
Keyword Spotting Competition, held during the 2016 International Conference on Frontiers in 
Handwriting Recognition. The training data of Botany was partitioned into three different 
training sets from smaller to larger (\textit{Train \RN{1}} 1684, \textit{Train \RN{2}} 5295, and \textit{Train \RN{3}} 21981 images). 
For our experiments we only use the \textit{Train \RN{3}} partition.
\subsection{Evaluation Protocol}\label{sec:protocol}
We evaluate the four CNN architectures for the data sets GW, IAM and Botany in the segmentation-based word spotting standard protocol 
proposed in \cite{Almazan14}. For the \gls{qbe} scenario all test images are considered as query which occur at least 
twice in the test set. For \gls{qbs} only unique string are used as queries. 
The CNNs predicte an attribute representation for each given query, afterwards a nearest neighbor search is performed 
by comparing the attribute vectors using the \textit{cosine similarity}. 
The Retieval lists are obtained  by sorting the list items by their cosine distances. 

We use the \emph{mean Average Precision (mAP)} as the performance metric by computing 
a \emph{interpolated Average Precision} for each query as recommended in \cite{Sudholt16}.

\subsection{Training Setup}\label{sec:training}
For training the different \glspl{cnn}, we use hyper parameters as suggested by \cite{Sudholt2018-ACF}:
All networks are trained using the \gls{bcel} and \gls{adam}~\cite{Kingma2015-AAM}.
As labels, we use \gls{phoc} vectors with levels 1, 2, 4, 8 and omit bi- or trigrams.
The momentum values $\beta_1$ for the mean and $\beta_2$ for the variance are set to the recommended $0.9$ and $0.999$ respectively while the variance flooring parameter $\epsilon$ is set to $10^{-8}$.
The initial learning rate is set to $10^{-4}$ and divided by $10$ after $70\,000$ training iterations for the \gls{gw} and Botany.
Training on these two data sets is carried out for a total of $80\,000$ iterations
For the \gls{iam} the step size is set to $100\,000$ and training is run for $240\,000$ iterations.

All networks are trained from scratch without the help of additional data.
As initialization, we follow the strategy proposed in \cite{He2015-DDI}:
For each layer $l$ the weights are sampled from a zero-mean normal distribution with variance $\frac{2}{n_l}$ where $n_l$ is the total number of trainable weights in layer $l$.

Similar to \cite{Sudholt2018-ACF}, the images are not preprocessed but the pixel values are scaled such that black pixels have a pixel value of $1.0$ while white pixels $0.0$.
We also augment the training set using the algorithm presented in \cite{Sudholt2018-ACF}.
This way, training images are generated such that there exists an equal amount of images per class used during training and the total amount of images is $500\,000$.

\footnotetext[1]{\footnotesize results obtained with additional annotated training data \cite{Wilkinson16}}

\subsection{Results \& Discussion}\label{sec:results}
Table \ref{tab:results} lists the results for the \gls{qbe} and \gls{qbs} 
experiments on three benchmarks. As expected the LeNet  
perform poorly on all three benchmarks, since this architecture has the least 
parameters and way too few feature maps. With only $70$ $(20+ 50)$ filter kernels 
in summation, the LeNet is not powerful enough to learn 
a rich feature representation. Even with more feature maps using 
only $500$ neurons in both hidden layers on the fully connected part, 
the \gls{mlp} would still perform poor. 
Nevertheless, with a relatively small 
\gls{cnn} like \gls{phoclenet} a mAP of $80.68\%$ for \gls{qbs} on the GW 
dataset is achieved. 
One can argue that word images from the GW dataset, written by 
the same writer, have very similar visual appearances, and 
thus a \gls{cnn} do not require an enormous amount of parameters 
to learn a mapping from visual to textual representations.
The \gls{tppnet} slightly outperforms the other three architectures on 
the GW dataset, but the \gls{phocresnet} achieves the best results on the IAM 
database. If we compare the \gls{tppnet} with the \gls{phocresnet}, we can 
argue that the GW dataset has less word instances and less word 
images. In contrast to the \gls{phoclenet}, the \gls{phocresnet} has 
to many parameters (over $144$ million) whereas the \gls{tppnet} 
contain $50$ million. 
Experiments showed, that even decreasing the depth of the \gls{tppnet} 
results in equal mAPs for \gls{qbe} and \gls{qbs} on the GW dataset. 
We assume that it is this over parameterization resulting in 
a degradation of a \gls{cnn} as mentioned in \cite{He16}. 

Suprisingly the \gls{phocdensenet} performs poor on both IAM and Botany.
Even though the PHOCDenseNet employs significantly more layers than the
TPP-PHOCNet and PHOCResNet, it is not able to achieve comparable
results.
We argue that the main disadvantage of the proposed architecture is the 
combination of a TPP layer and a \gls{mlp}.
While the TPP-PHOCNet contains $512$ feature-maps in its final
convolutional layer, the PHOCDenseNet has $916$.
Although an increased number of feature-maps are generated, we argue that
their representational power is comparable low.
As discussed in \cite{Huang2017}, the classification layer of a
\gls{densenet} heavily focuses on the complex feature-maps created late in
the network.
Pre-experimental evaluations have shown that increasing the number of
feature-maps, by simply adding layers to the network, does not increase the
mAP nether in \gls{qbe} nor in \gls{qbs}.
An increased number of feature-maps also corresponds to a heavy increase
of parameters in the \gls{mlp}.
We believe, this overparameterization prevents the network of benefiting
from an increased network deep.


\section{Conclusion}\label{sec:conclusion}

In this work, we explored four \gls{cnn} architectures 
on three standard word spotting benchmarks for 
Query-by-Example and Query-by-String. 
The experiments conducted show that 
deeper \gls{cnn} architectures do not 
necessarely perform better on word spotting 
tasks. On the contrary, the recently proposed 
\gls{densenet} \cite{Huang2017} performs the 
worst on all three benchmarks excluding the 
\gls{phoclenet}. This leads us to the 
hyperthesis that nether increasing the depth nor 
the number of parameters in the convolutional 
part achieves significantly higher performance 
in word spotting. Instead, future research 
needs to focus on word embeddings incorporating 
more informations than only character occurence 
or position.

\bibliographystyle{IEEEtran}
\bibliography{literature}
\end{NoHyper}
\end{document}